\documentclass[sn-mathphys,Numbered]{sn-jnl}

\usepackage{graphicx}%
\usepackage{multirow}%
\usepackage{amsmath,amssymb,amsfonts}%
\usepackage{amsthm}%
\usepackage{mathrsfs}%
\usepackage[title]{appendix}%
\usepackage{xcolor}%
\usepackage{textcomp}%
\usepackage{manyfoot}%
\usepackage{booktabs}%
\usepackage{algorithm}%
\usepackage{algorithmicx}%
\usepackage{algpseudocode}%
\usepackage{listings}%
\usepackage{subfigure}
\usepackage[section]{placeins}

\theoremstyle{thmstyleone}%
\theoremstyle{thmstyletwo}%

\theoremstyle{thmstylethree}%

\begin{document}

\title[Article Title]{Skipped Feature Pyramid Network with Grid Anchor for Object Detection}


\author[1]{\fnm{Li} \sur{Pengfei}}\email{lpf722@163.com}

\author[1]{\fnm{Wei} \sur{Wei}}\email{510320796@qq.com}

\author[2,3]{\fnm{Yan} \sur{Yu}}\email{yy@hzmbo.com}

\author*[4]{\fnm{Zhu} \sur{Rong}}\email{zhurong@whu.edu.cn}

\author[4]{\fnm{Zhou} \sur{Liguo}}\email{zhouliguo@whu.edu.cn}

\affil[1]{\orgname{Wuhan Maritime Communication Research Institute}, \orgaddress{\city{Wuhan}, \country{China}}}

\affil[2]{\orgname{Chongqing Jiaotong University}, \orgaddress{\city{Chongqing}, \country{China}}}

\affil[3]{\orgname{Hong Kong-Zhuhai-Macao Bridge Authority}, \orgaddress{\city{Zhuhai}, \country{China}}}

\affil[4]{\orgname{Wuhan University}, \orgaddress{\city{Wuhan}, \country{China}}}


\abstract{CNN-based object detection methods have achieved significant progress in recent years. The classic structures of CNNs produce pyramid-like feature maps due to the pooling or other re-scale operations. The feature maps in different levels of the feature pyramid are used to detect objects with different scales. For more accurate object detection, the highest-level feature, which has the lowest resolution and contains the strongest semantics, is up-scaled and connected with the lower-level features to enhance the semantics in the lower-level features. However, the classic mode of feature connection combines the feature of lower-level with all the features above it, which may result in semantics degradation. In this paper, we propose a skipped connection to obtain stronger semantics at each level of the feature pyramid. In our method, the lower-level feature only connects with the feature at the highest level, making it more reasonable that each level is responsible for detecting objects with fixed scales. In addition, we simplify the generation of anchor for bounding box regression, which can further improve the accuracy of object detection. The experiments on the MS COCO and Wider Face demonstrate that our method outperforms the state-of-the-art methods.}

\keywords{CNN, Feature Pyramid Network, Object Detection, Face Detection}



\maketitle

\section{Introduction}\label{sec1}

Object detection is an important task in computer vision and has been widely studied in the past decades. Nowadays, many emerging applications, such as autonomous driving~\cite{nrp, sim} and face recognition~\cite{ml1, ml2}, hinge on object detection. In recent years, Convolutional Neural Networks (CNN)~\cite{alexnet} have achieved significant progress in object detection. An object detection network can usually be divided into two parts. One part is used to extract features, and the other part is used to predict the objects' bounding boxes. The performance of object detection relies on better feature extraction. For improving the performance of CNN in feature extracting, more and more complicated components are proposed. However, the most popular networks always have a similar structure to the very original LeNet~\cite{lenet}, which contains several pooling operations and produces pyramid-like feature maps. In the feature pyramid, the higher-level features, which have lower-resolution, have stronger semantics, while the lower-level features, which have higher-resolution, have weaker semantics. R-CNN series~\cite{rcnn, fast, faster} use the highest-level feature, which has the strongest semantics, to detect objects. For detecting multi-scale objects accurately, the image is resized to different sizes and then input the network repeatedly, which increases the time cost. The SSD-based methods~\cite{ssd} use features of multi-levels to detect objects, and this method is more effective for detecting multi-scale objects.

\begin{figure}[H]
\begin{center}

\subfigure[FPN]{
\includegraphics[width=0.47\textwidth]{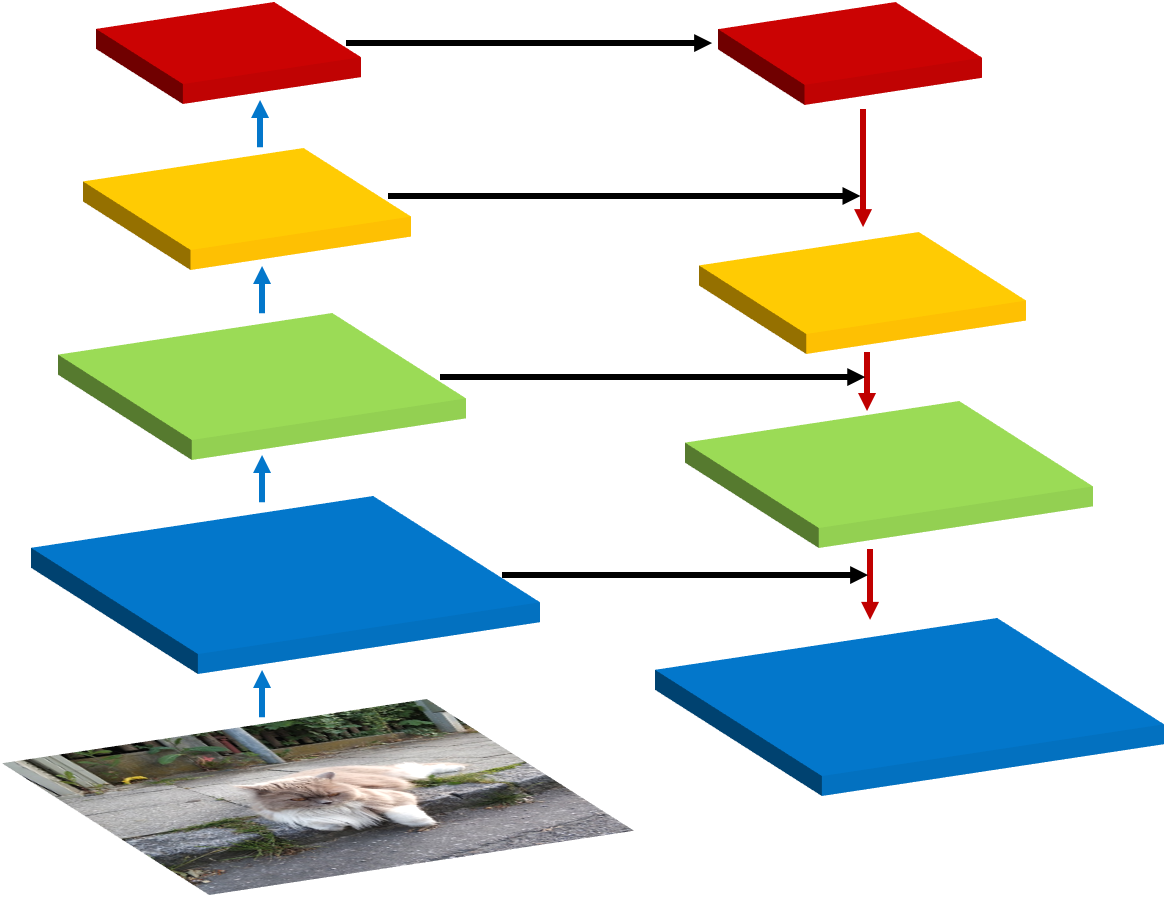}
}
\subfigure[SFPN]{
\includegraphics[width=0.47\textwidth]{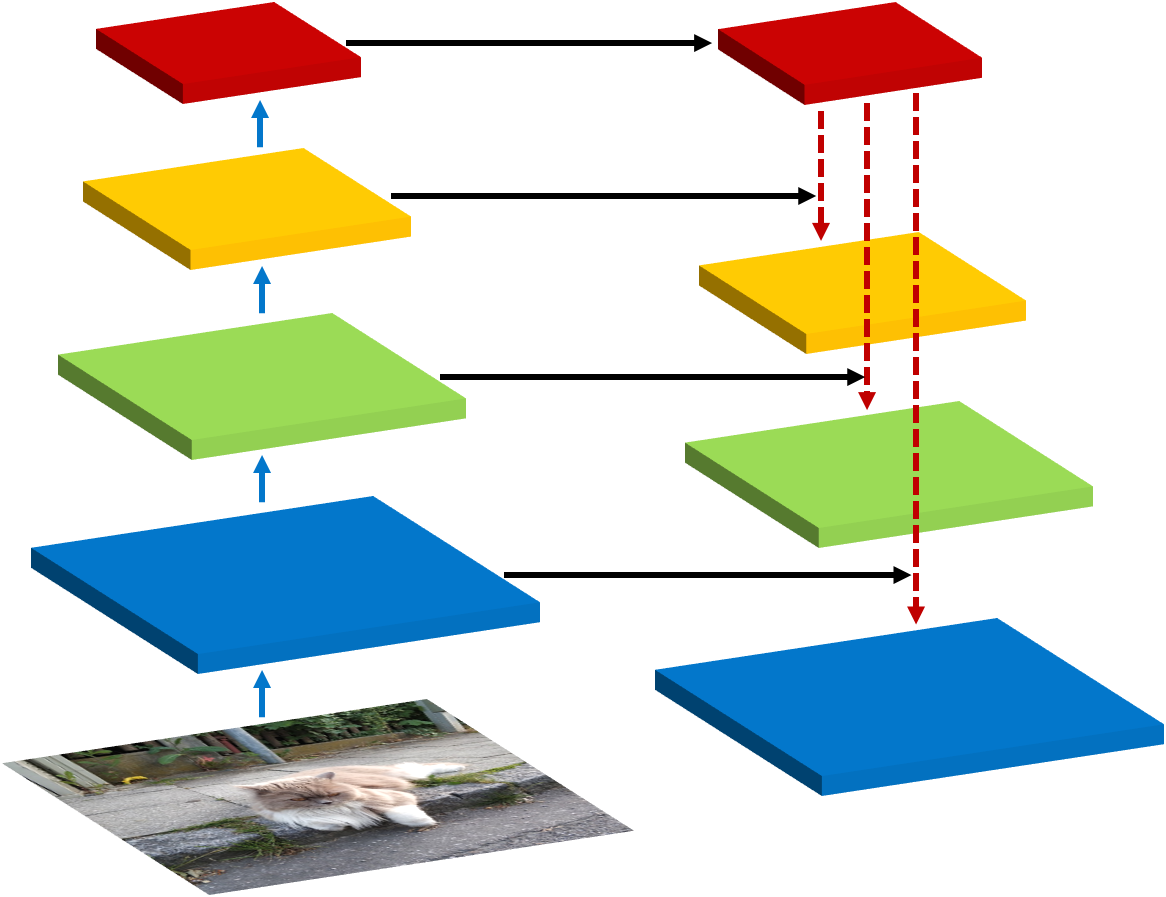}
}
\end{center}
   \caption{Comparison between FPN and Skipped FPN (SFPN).}
\label{figure 1}
\end{figure}

In order to further improve the accuracy of object detection, Feature Pyramid Network (FPN)~\cite{fpn} is introduced to enhance the semantics of features at all levels. As shown in Figure~\ref{figure 1}(a), FPN up-scales the feature of highest-level and fuses it with the feature of the lower-level. Then the fusion result is up-scaled to fuse with the feature of more lower-level. All the needed low-level features are fused with the feature of the highest-level and obtain stronger semantics. The Path Aggregation Network (PAN)~\cite{pan} extents FPN by fusing the features again from bottom to top. In PAN, the lowest-level feature of FPN is down-scaled to fuse with the features of higher-levels. Although PAN is just a reverse operation of FPN, it can also enhance the semantics in the features which are used to detect multi-scale objects.

In this paper, we modify the FPN by reconnecting features. Figure~\ref{figure 1} shows the difference between FPN and our method. We connect the feature of highest-level with the features of lower-levels directly. The feature of each lower-level is fused with the feature of highest-level, but has no relationship with other levels. The semantics of the highest-level can be preserved to the greatest extent. In addition, the features with different resolutions are supposed to detecting objects of different sizes. The points in the features of lower-levels have smaller Receptive Field~\cite{rf}. It is suitable to use lower features to detect smaller objects. The points in the features of higher-levels have larger Receptive Field, so the higher features are dedicated to detecting larger objects. Our method can reduce mutual interference by cutting the relationship between the features of different levels. In addition, we modify the anchor generation and bounding-box regression to get a better accuracy of object detection. We test our method on MS COCO~\cite{coco} and Wider Face~\cite{wider}. The experiments demonstrate that our method outperforms the state-of-the-art methods.

\section{Related Works}

\subsection{Object detection}

Since the R-CNN~\cite{rcnn} succeeded in object detection, various excellent object detection algorithms have emerged one after another. Object detection algorithms can be broadly divided into two categories: one-stage and two-stage. R-CNN and its improved versions, Fast R-CNN~\cite{fast} and Faster R-CNN~\cite{faster}, all belong to the two-stage. They firstly search the region proposals and then classify the contents in the proposals. SSD~\cite{ssd} and YOLO~\cite{yolo} are two representative one-stage methods. They are based on global regression or classification in which the image pixels are mapped directly to bounding box coordinates and class probabilities. The two-stage methods are more time-consuming than the one-stage methods because of the existence of the region proposal part. Both one-stage and two-stage methods contain bounding-box regression for accurate detection.

\subsection{Face detection}

As a special object detection task, the progress of face detection benefits from the development of object detection. Many CNN-based object detection methods are used in face detection. For improving the accuracy on the popular face benchmarks~\cite{fddb, wider}, loads of complicated components are added to the networks, which result in the degradation of speed. PyramidBox~\cite{pyramidbox} proposes a context-assisted single-shot face detector. DSFD~\cite{dsfd} introduces a feature enhance module to extend the single-shot detector~\cite{ssd} to the dual shot detector. EXTD~\cite{extd} generates the feature maps by iteratively reusing a shared lightweight and shallow backbone network instead of a single backbone network. These SSD-based~\cite{ssd} face detectors all get high accuracy on the Wider Face benchmark. Based on the YOLOv5~\cite{yolov5} object detector, YOLO5Face~\cite{yolo5face} achieves a better performance on the Wider Face benchmark by adding a five-point landmark regression head and using the Wing loss function~\cite{wing}. 

\section{Method}

\begin{figure*}[h]
\begin{center}
\includegraphics[width=\textwidth]{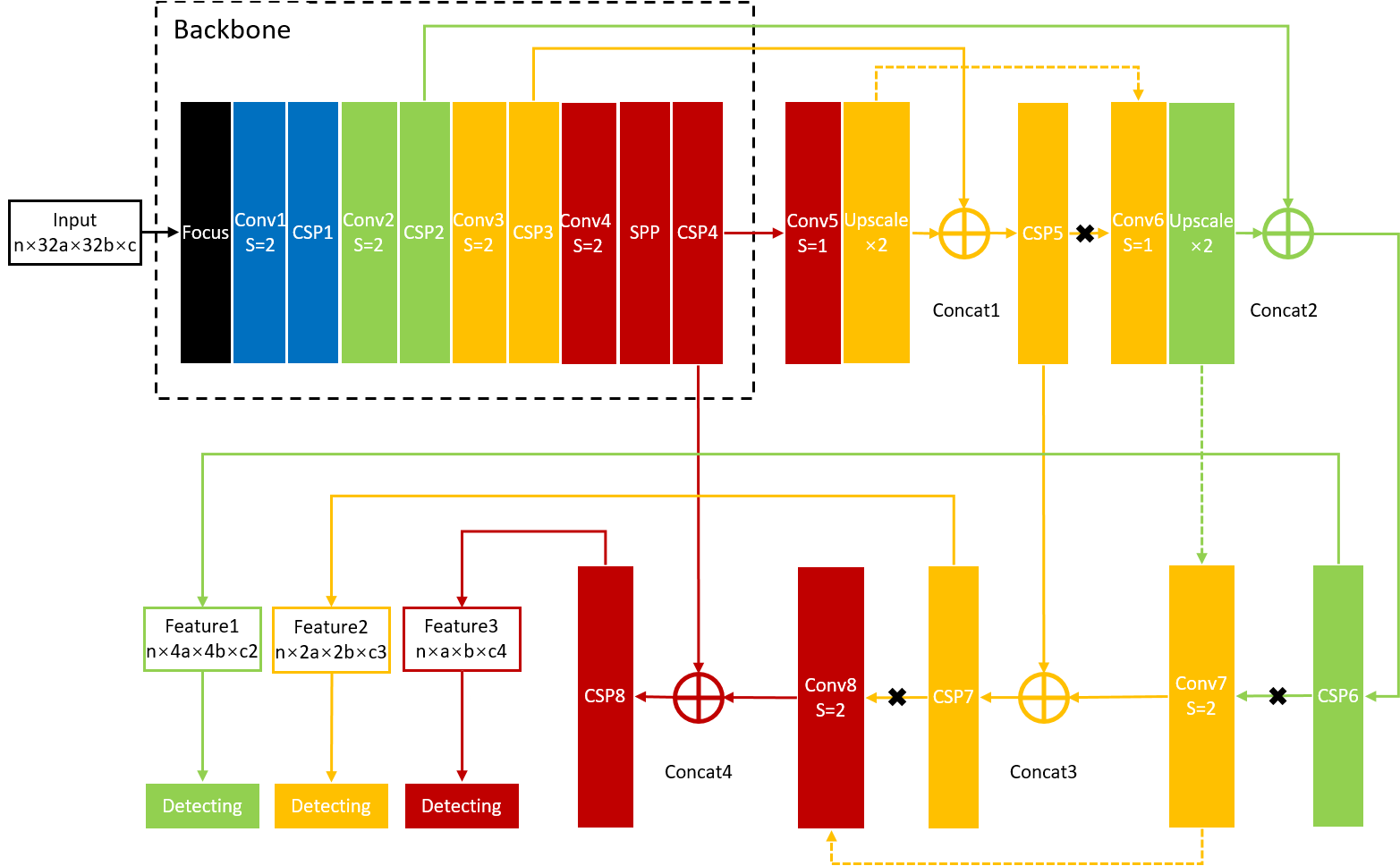}
\end{center}
   \caption{The object detection network with Skipped FPN (SFPN). The \includegraphics[width=0.2cm]{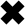} indicates the connections are cut off and the dotted arrow lines are the new connections.}
\label{figure 2}
\end{figure*}


\subsection{Network}

We use CSPNet~\cite{csp} in YOLOv5~\cite{yolov5} as the base framework of our method. Our network is shown in Fig.~\ref{figure 2}. The network outputs three feature maps for object detection. The feature extracting part includes a backbone, FPN~\cite{fpn}, and PAN~\cite{pan}. The Focus~\cite{focus}, CSP~\cite{csp}, and SPP~\cite{spp} are introduced for better feature extraction. The \textit{Conv with S=2} and \textit{Upscale} operation layers downscale the feature map to 1/4 and enlarge the feature map to 4 times, respectively. The operation layers with the same color output feature maps with the same size. At the end of the network, there are three branches for detecting objects. The widths and heights of Feature1-3 are 1/8, 1/16, and 1/32 of the width and height of the input image.

According to the description of SFPN in Figure~\ref{figure 1}(b), we cut some connections and rebuild some connections as the \includegraphics[width=0.2cm]{fig/x.png} and dotted arrow lines indicate in Figure~\ref{figure 2}.
\begin{figure}[h]
\begin{center}

\subfigure[Image and Its Feature Map Pyramid Produced by CNN.]{
\includegraphics[width=8.6cm]{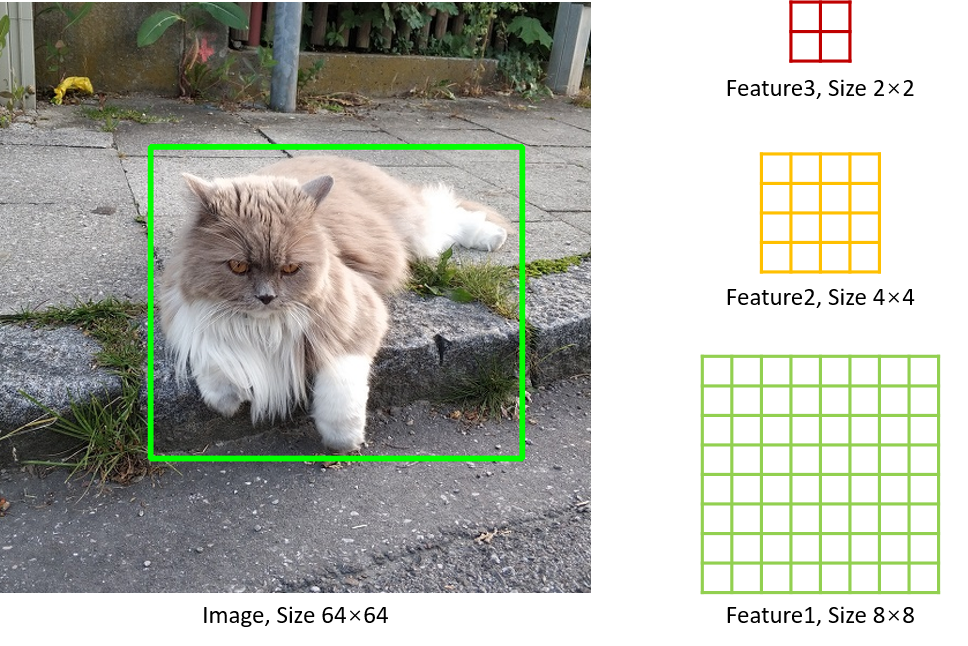}
}
\subfigure[The White Squares are the Anchors Selected According to the Feature Maps]{
\includegraphics[width=\textwidth]{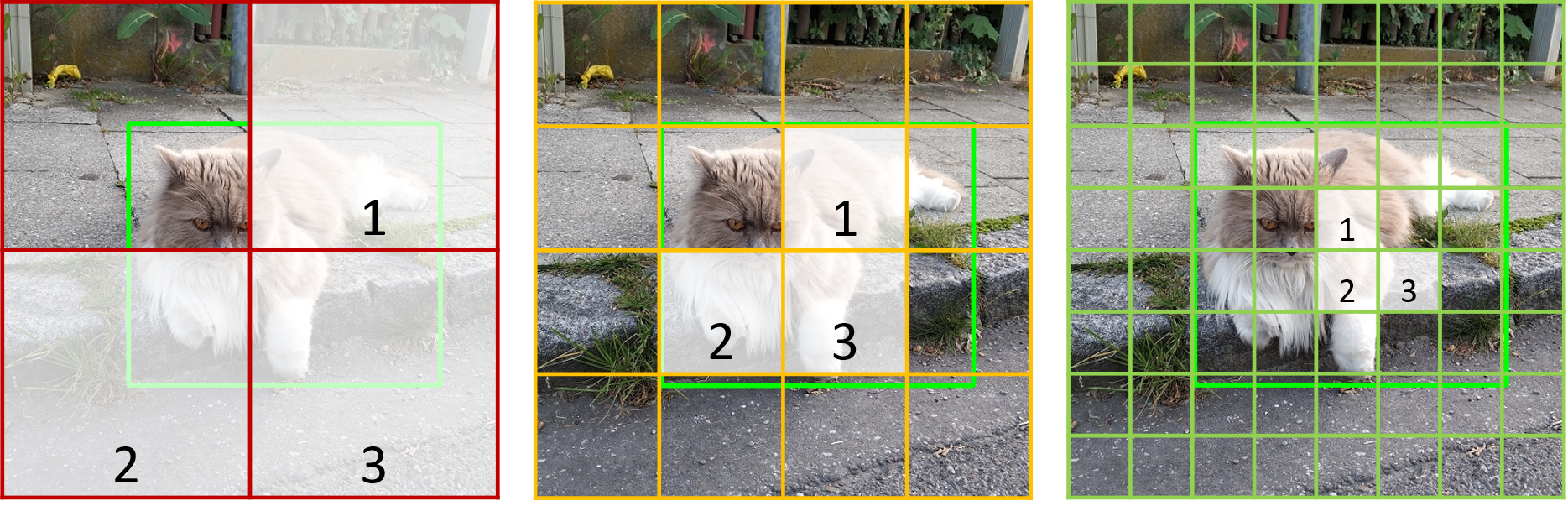}
}
\end{center}
   \caption{(a) An image and its feature map pyramid produced by CNN. The green box is the ground-truth bounding-box. (b) The image is divided into grid images according to the sizes of the feature maps. In each grid image, one grid that includes the center point of the ground-truth and two grids closest to the center point of the ground-truth are selected as the anchors.}
\label{figure 3}
\end{figure}

\subsection{Grid Anchor}

We first divide the input image into grids according to the sizes of the feature maps. As shown in Fig.~\ref{figure 3}(b), if the size of input image is $w\times h$, Feature1-3 divide the image into four grid images with $w/8\times h/8$, $w/16\times h/16$, and $w/32\times h/32$ grids. In each grid image, the three white grids are selected as the anchors. One includes the center point of the ground-truth, and two are closest to the center point of the ground-truth. For Feature1-3, the sizes of the anchors are $8\times 8$, $16\times 16$, and $32\times 32$. If the anchor size is $d\times d$ in one branch, the ground-truth whose longer side is longer than $d^2$ should be ignored when selecting anchors.

\subsection{Bounding-box Regression}

Like YOLOv5~\cite{yolov5}, we code the ground-truth bounding-box first. In pixel domain of an image, the ground-truth bounding-box can be denoted as $(t_x, t_y, t_w, t_h)$ and the anchor is $(g_x, g_y, d, d), d\in\{8,16,32\}$. $(t_x, t_y)$ is the center point of the ground-truth and $(g_x, g_y)$ is the upper-left corner of the anchor. $(t_w, t_h)$ and $(d, d)$ are the widths and heights. We code the ground-truth to $T_c(t_{x\_c},t_{y\_c},t_{w\_c}, t_{h\_c})$ by (1).
\begin{equation}
\begin{split}
&t_{x\_c}=\frac{t_x-g_x}{d}\\
&t_{y\_c}=\frac{t_y-g_y}{d}\\
&t_{w\_c}=\frac{t_w}{d}\\
&t_{h\_c}=\frac{t_h}{d}
\end{split}
\end{equation}
The $(t_{x\_c},t_{y\_c})$ is the relative position to the upper-left corner of the anchor in the feature domain and the values are between -0.5 and 1.5. $(t_{w\_c}, t_{h\_c})$ is the width and height of the ground-truth in the feature domain.

At the end of the network, the channel of the feature map is transformed to 5. If a feature point, which is a 5-dimension vector $(z_0, z_1, z_2, z_3, z_4)$, corresponds to a selected anchor, it is used for predicting the bounding-box. The vector should be divided into two parts. The first four are transformed to denote the predicted coded bounding-box. The last one represents the score of the bounding-box.  We first normalize the values to between 0 and 1 by Sigmoid (2), and the result is $(z_0’, z_1’, z_2’, z_3’, z_4’)$.
\begin{equation}
z' = \frac{1}{1+e^{-z}}
\end{equation}
The predicted coded bounding-box $P_c(p_{x\_c}, p_{y\_c}, p_{w\_c}, p_{h\_c})$ can be get by (3).
\begin{equation}
\begin{split}
&p_{x\_c}=2z_0'-0.5\\
&p_{y\_c}=2z_1'-0.5\\
&p_{w\_c}=\frac{d^{2z_2'}}{d}\\
&p_{h\_c}=\frac{d^{2z_3'}}{d}
\end{split}
\end{equation}
We use CIoU~\cite{ciou} to measure the predicted bounding-box. The regression loss of each branch in the network is defined in (4),
\begin{equation}
L_{reg}^{(d)}=\frac{1}{N^{(d)}}\sum_{n=1}^{N^{(d)}}1-CIoU[(P{_c}^{(n)},T{_c}^{(n)})]
\end{equation}
where $N^{(d)}$ is the number of anchors that are selected for bounding-box regression in one batch. 
The regression loss of the whole network is (5).
\begin{equation}
L_{reg}=L_{reg}^{(8)}+L_{reg}^{(16)}+L_{reg}^{(32)}
\end{equation}
The score of the bounding-box can be represented by CIoU. The score loss of each branch in the network is defined in (6).
\begin{equation}
\begin{aligned}
L_{score}^{(d)} = -\frac{1}{N^{(d)}}\sum_{n=1}^{N^{(d)}}[(1-CIoU^{(n)})\log(1-z_{4}^{'(n)})\\+CIoU^{(n)}\log(z_{4}^{'(n)})]
\end{aligned}
\end{equation}
The score loss of the whole network is 
\begin{equation}
L_{score}=L_{score}^{(8)}+L_{score}^{(16)}+L_{score}^{(32)}
\end{equation}
The loss function of the whole network is (8),
\begin{equation}
L=L_{score}+L_{reg}+\lambda ||W||^2
\end{equation}
where $W$ are the weights in the network and  $\lambda ||W||^2$ is added to avoid over-fitting.
In testing, the predicted bounding-box $(p_x, p_y, p_w, p_h)$ can be obtained by (9).
\begin{equation}
\begin{split}
&p_{x}=(2z_0'-0.5)'d+g_x\\
&p_{y}=(2z_1'-0.5)'d+g_y\\
&p_{w}=d^{2z_2'}\\
&p_{h}=d^{2z_3'}
\end{split}
\end{equation}

\subsection{Post Processing}

In testing, if an image is detected as containing objects, each branch will output several predicted bounding-boxes for each object. We take three steps to select the most appropriate bounding-boxes. Firstly, we filter out the over-large bounding-boxes, whose longer side-lengths are longer than $d^2$, in the outputs of the branches of Feature1-2. Secondly, we use the Non-Maximum Suppression (NMS) to select the best bounding-box in each branch. Finally, we use NMS again to select the best bounding-box among all branches.

\section{Experiment}

\begin{figure}[h]
\begin{center}
\subfigure[MS COCO]{
\includegraphics[height=4.1cm]{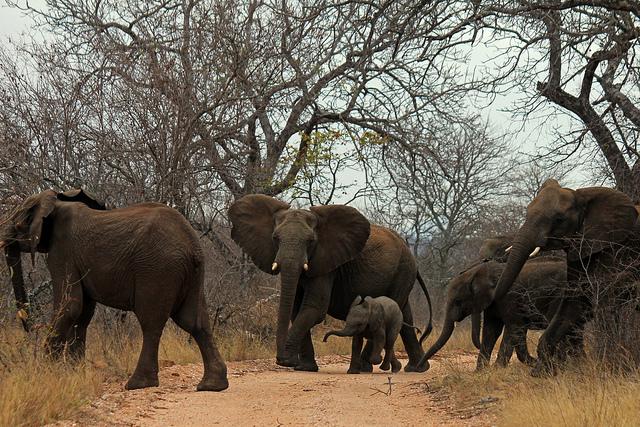}
}
\subfigure[Wider Face]{
\includegraphics[height=4.1cm]{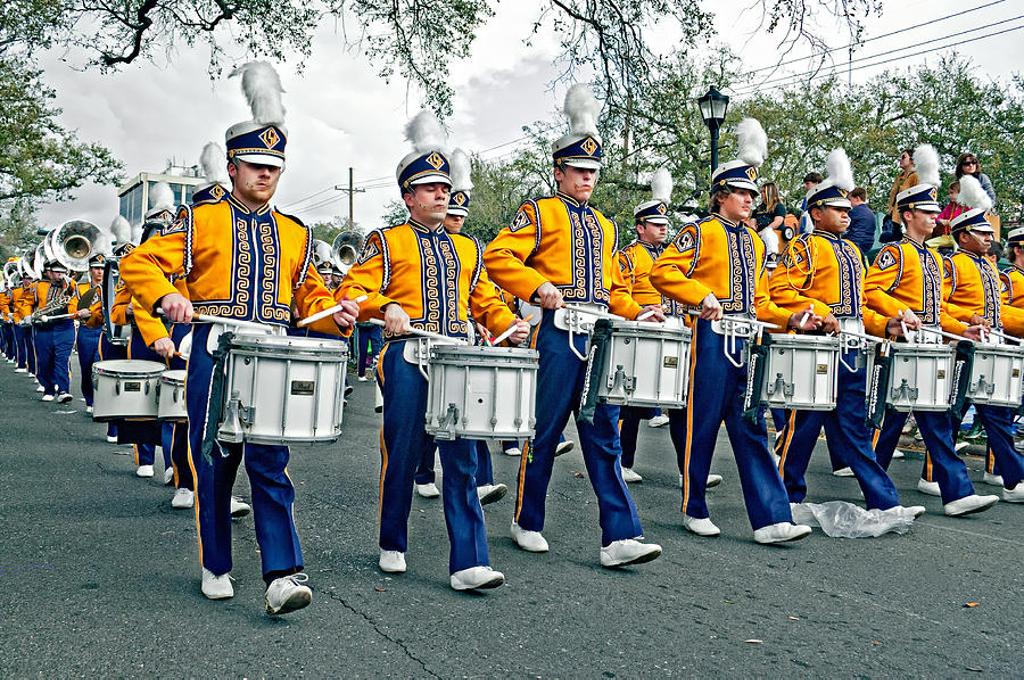}
}
\end{center}
   \caption{Examples in MS COCO and Wider Face.}
\label{figure 4}
\end{figure}

We train and test our method on MS COCO and Wider Face benchmarks to compare with the state-of-the-art methods. The MS COCO dataset contains more than 200,000 images and 80 object categories which are split into three subsets train, validation, and test. All object instances are annotated with a bounding box. Annotations on the training and validation sets (with over 500,000 objects) are publicly available. Figure~\ref{figure 4}(a) shows a sample of MS COCO. The WIDER FACE contains 32,203 images and 393,703 labeled faces with a high degree of variability in scale, pose, and occlusion, as shown in Figure~\ref{figure 4}(b). The whole dataset is divided into training, validation, and testing sets. The validation and testing sets are split into three subsets (easy, medium, and hard). 

The network is trained on Nvidia V100 GPUs with PyTorch~\cite{pytorch}. The hyper-parameters of the backbone are the same as the YOLOv5x~\cite{yolov5}. The hyper-parameters of the other part are determined by the backbone. For data preprocessing, we normalize the pixel values to [0,1] and use Mosaic data augmentation~\cite{yolov4}. The size of the image is randomly re-scaled from 0.5 to 1.5. The batch stochastic gradient descent~\cite{lenet} is used in training.

\begin{table}[]
\renewcommand\arraystretch{1.5}
    \centering
    \begin{tabular}{c|c|c|c|c|c}
    \hline
    \hline
        Model & Size & mAP val 50-95 & mAP val 50 & Params (M) &FLOPs @640 (B)\\
        \hline
        YOLOv5x & 640 & 50.7  & 68.9 & 86.7 & 205.7\\
        YOLOv5xSfpn & 640 & 51.7  & 70.2 & 86.7 & 205.7\\
        YOLOv5xSfpnGa & 640 & 52.0  & 70.8 & 86.7 & 205.7\\
    \hline
    \hline
    \end{tabular}
    \caption{Comparison of Object Detection Results on MS COCO.}
    \label{tab:my_label}
\end{table}

Table~\ref{tab:my_label} shows the comparison between our method and YOLOv5x original method. YOLOv5Sfpn is the modified YOLOv5 with SFPN, and YOLOv5SfpnGa is YOLOv5Sfpn with grid anchor. The results show that SFPN can improve the Mean Average Precision (mAP) of object detection and grid anchor can further improve mAP slightly.

\begin{figure*}[h]
\begin{center}
\subfigure[Easy-Val]{
\includegraphics[width=6.2cm]{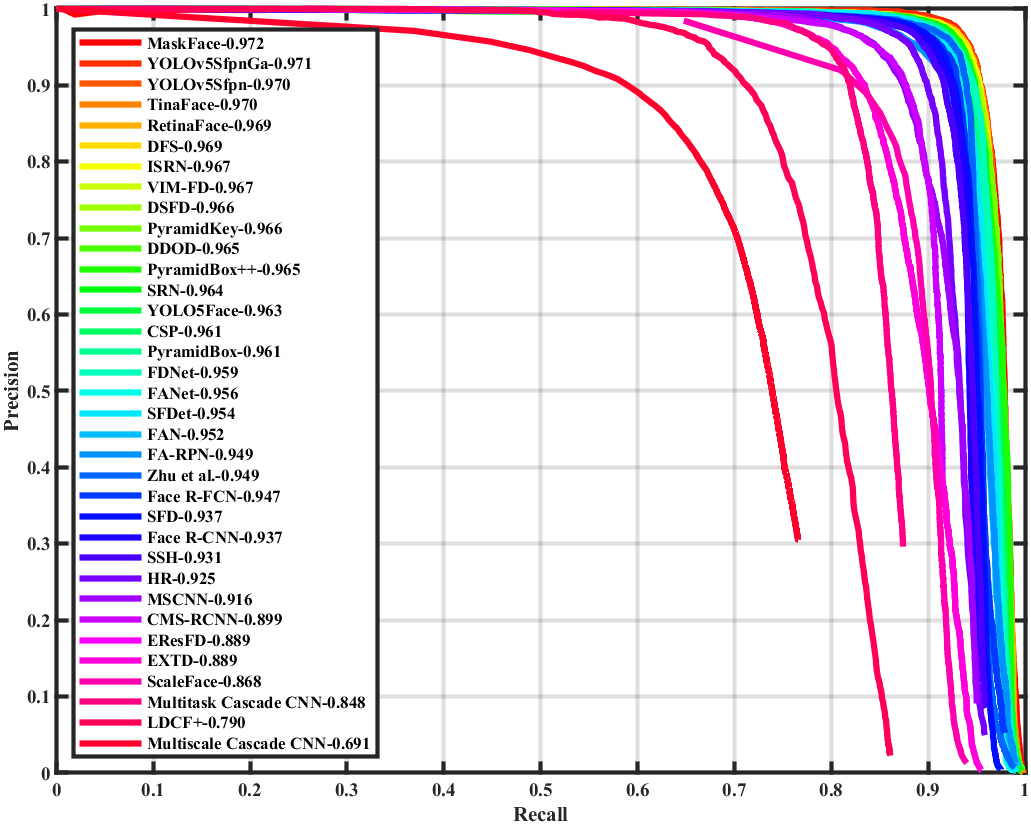}
}
\subfigure[Easy-Test]{
\includegraphics[width=6.2cm]{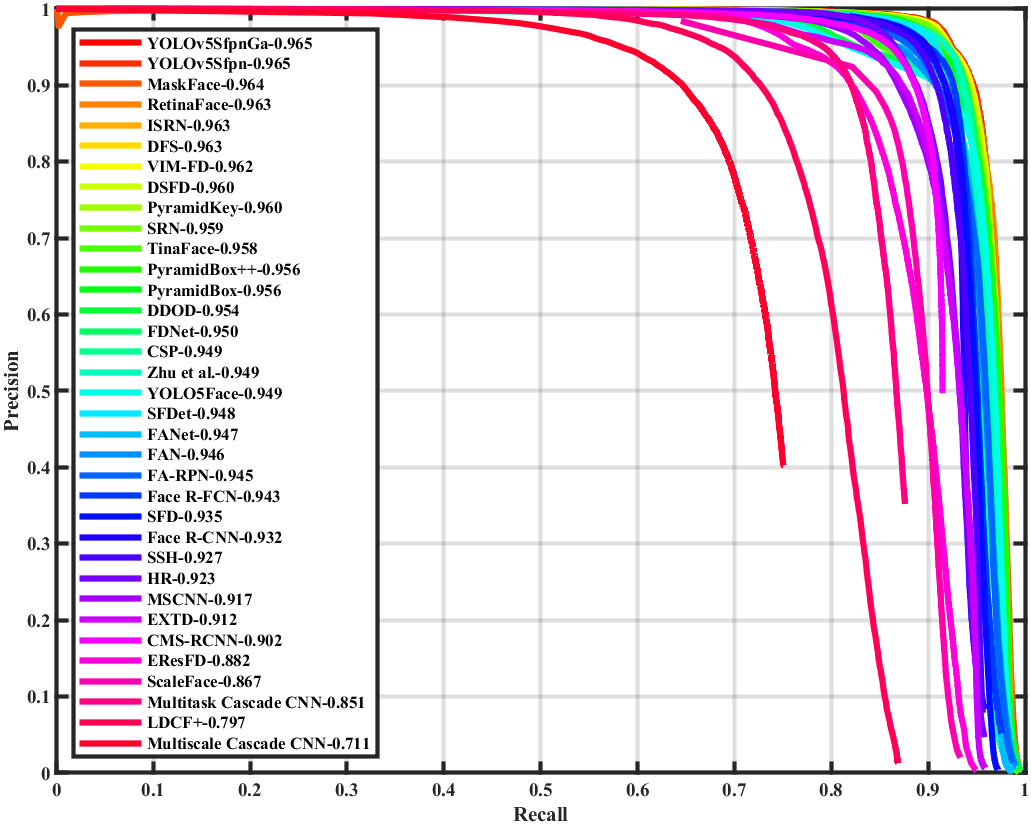}
}
\subfigure[Medium-Val]{
\includegraphics[width=6.2cm]{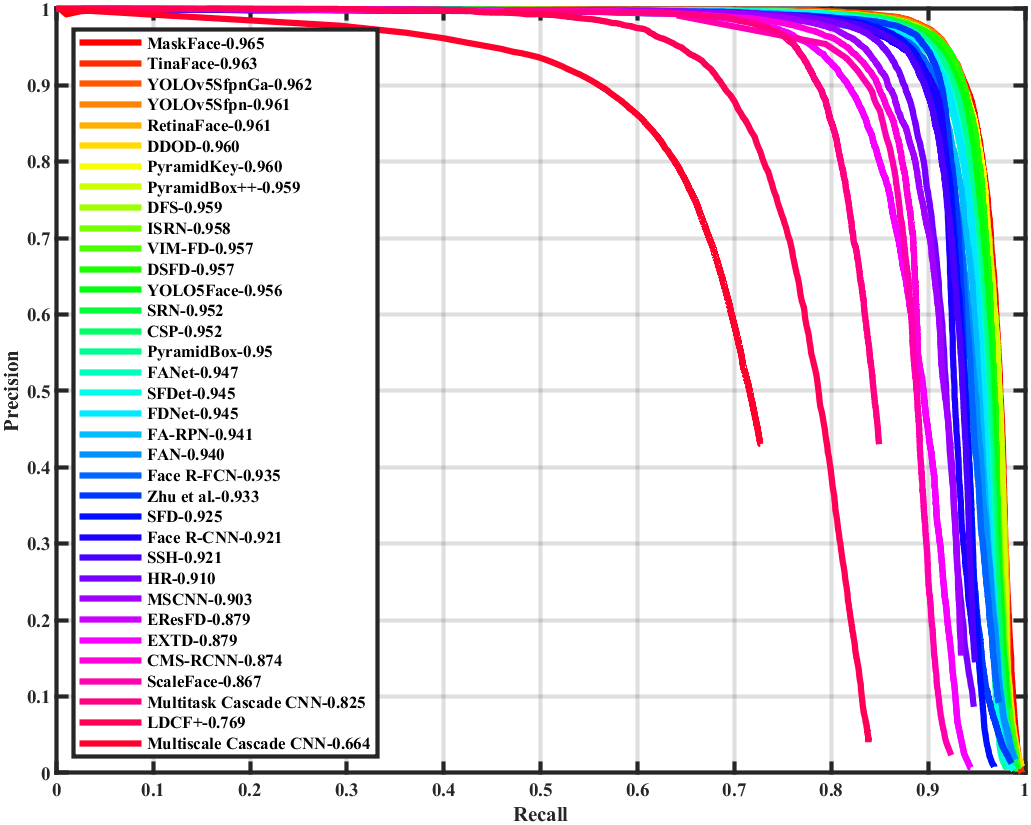}
}
\subfigure[Medium-Test]{
\includegraphics[width=6.2cm]{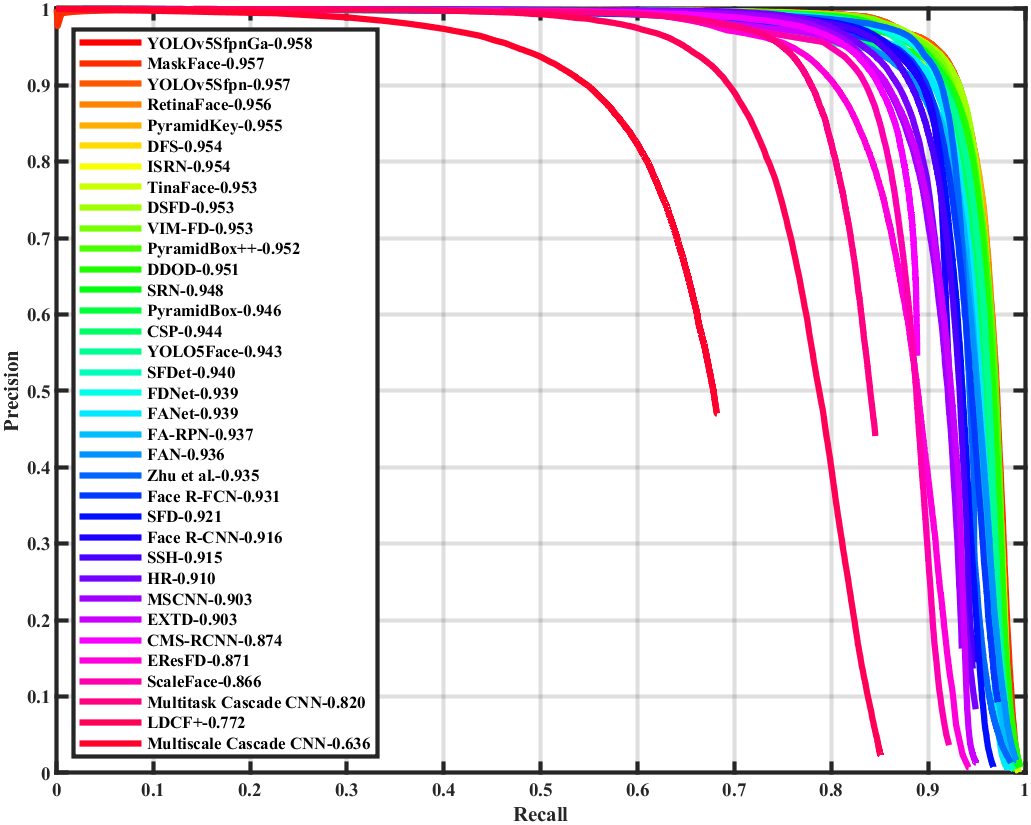}
}
\subfigure[Hard-Val]{
\includegraphics[width=6.2cm]{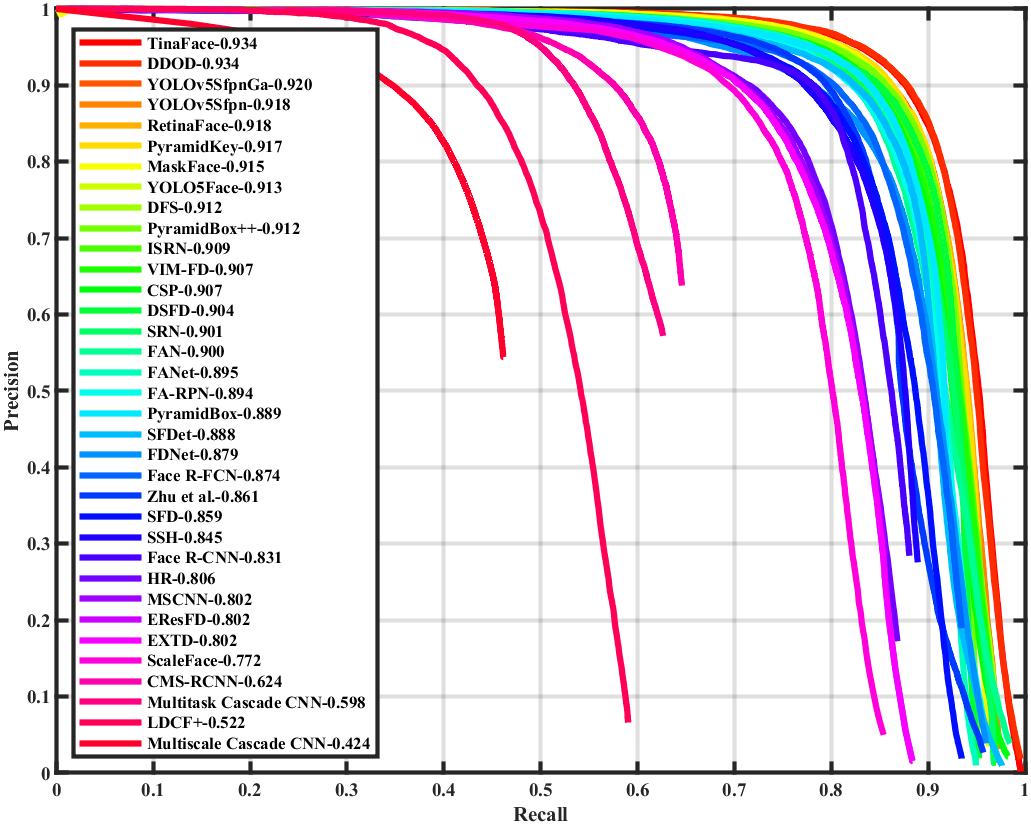}
}
\subfigure[Hard-Test]{
\includegraphics[width=6.2cm]{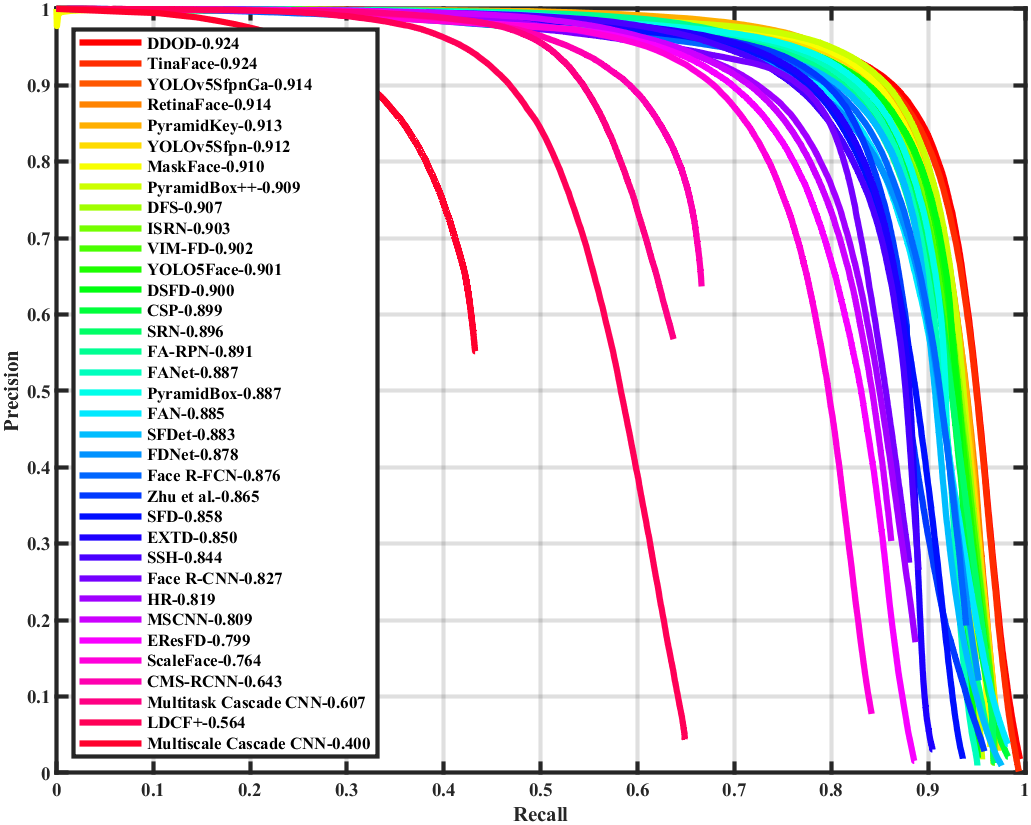}
}
\end{center}
   \caption{PR Curves on the Validation and Testing Sets of Wider Face.}
\label{figure 5}
\end{figure*}

Wider Face releases the results of state-of-the-art methods. Figure~\ref{figure 5} shows the comparison between our method and state-of-the-art. YOLO5Face~\cite{yolo5face} uses YOLOv5x's method to do face detection. In Figure~\ref{figure 5}, YOLOv5Sfpn outperforms YOLO5Face~\cite{yolo5face} and YOLOv5SfpnGa outperforms YOLOv5Sfpn. Both YOLOv5Sfpn and YOLOv5SfpnGa reach the state-of-the-art and even outperform the state-of-the-art in some subsets. 

\section{Conclusion}

CNN produces a feature pyramid, and features in different levels have Receptive Field with different sizes. The features in different levels are used to detect faces with different scales. To enhance the semantics of the lower-level features, the highest-level feature is fused into the lower-level features. In this paper, we change the FPN to SFPN by connecting lower-level features with the highest-level feature directly. This method can enhance the semantics of the lower-level features and reduce the mutual influence between the lower-levels. Each level of the feature pyramid is more dedicated to detecting objects with sizes corresponding to the resolution of the feature. By combining the grid anchor, the Skipped FPN can further improve the performance of object detection.

\section{Acknowledgement}

This research is funded by the National Key Research and Development Program of China (2019YFB1600700).


\bibliography{sn-bibliography}

\end{document}